\documentclass[sigconf,authorversion,nonacm]{acmart}
\usepackage{caption}
\usepackage{subcaption}
\usepackage{array}
\usepackage{hyperref, cleveref}
\usepackage[inline]{enumitem}
\usepackage{soul}
\usepackage{amsmath}
\AtBeginDocument{%
  \providecommand\BibTeX{{%
    \normalfont B\kern-0.5em{\scshape i\kern-0.25em b}\kern-0.8em\TeX}}}

\setcopyright{acmcopyright}
\copyrightyear{2023}
\acmYear{2023}
\acmDOI{XXXXXXX.XXXXXXX}

\acmPrice{15.00}
\acmISBN{978-1-4503-XXXX-X/18/06}

\begin{document}

\title{Machine Learning-Enhanced Prediction of Surface Smoothness for Inertial Confinement Fusion Target Polishing Using Limited Data}

\author{Antonios Alexos}
\affiliation{%
  \institution{University of California, Irvine}
  \country{United States of America}
}

\author{Junze Liu}
\affiliation{%
  \institution{University of California, Irvine}
  \country{United States of America}
}

\author{Akash Tiwari}
\affiliation{%
  \institution{Texas A\&M University}
  \country{United States of America}
}

\author{Kshitij Bhardwaj}
\affiliation{%
  \institution{Lawrence Livermore National Laboratory}
  \country{United States of America}
}

\author{Sean Hayes}
\affiliation{%
  \institution{Lawrence Livermore National Laboratory}
  \country{United States of America}
}

\author{Pierre Baldi}
\affiliation{%
  \institution{University of California, Irvine}
  \country{United States of America}
}

\author{Satish Bukkapatnam}
\affiliation{%
  \institution{Texas A\&M University}
  \country{United States of America}
}

\author{Suhas Bhandarkar}
\affiliation{%
  \institution{Lawrence Livermore National Laboratory}
  \country{United States of America}
}


\begin{abstract}
In Inertial Confinement Fusion (ICF) process, roughly a 2mm spherical shell made of high density carbon is used as target for laser beams, which compress and heat it to energy levels needed for high fusion yield. These shells are polished meticulously to meet the standards for a fusion shot. However, the polishing of these shells involves multiple stages, with each stage taking several hours. To make sure that the polishing process is advancing in the right direction, we are able to measure the shell surface roughness. This measurement, however, is very labor-intensive, time-consuming, and requires a human operator. We propose to use machine learning models that can predict surface roughness based on the data collected from a vibration sensor that is connected to the polisher. Such models can generate surface roughness of the shells in real-time, allowing the operator to make any necessary changes to the polishing for optimal result.
\end{abstract}







\maketitle

\section{Introduction}
\label{sec:intro}

Nuclear fusion as a source of energy is beginning to show promise as a powerful means to address energy crisis problem. Decades of sustained research on laser mediated nuclear fusion research has recently culminated in the demonstration of ignition. At the National Ignition Facility (NIF), in December 2022, an experiment yielded about 1.5 times greater fusion energy than the optical energy, which has generated much interest in developing nuclear fusion as a highly concentrated energy source that produces no harmful byproducts.

The approach used at NIF for this demonstration is called Inertial Confinement Fusion (ICF).  Here, a roughly 2mm spherical shell made of high density carbon (HDC), which is akin to diamond \cite{biener2009diamond}, houses the deuterium tritium fuel on the interior required for ICF.
The fusion process is initiated by 192 laser beams which imparts the energy to the shell causing the HDC shell to implode rapidly towards the fuel core which then compresses and heats it to levels ($10^9$ bars and $10^8$ K) needed to overcome the Coulombic repulsion between the deuterium and tritium nuclei.
In December 2022, the energy supplied by the laser was 2.05 MJ and the nuclear yield obtained was 3.15 MJ, demonstrating a sustained fusion reaction or ignition for the first time.

For a high yield, the surface quality of the shell (smooth, round, defect free, and concentric) has been identified as one of the major drivers which can be controlled during the fabrication process.
Such high surface quality requirement of the shells imposes stringent criteria for their finishing processes.
After these shell targets progress through the fabrication process, the shells that meet the surface quality standards become qualified for a fusion shot. 

The total time spent on finishing the shells to the required standards is exceedingly time-consuming, which stands as the main motivator for this work. We seek to investigate this process and employ advanced machine learning techniques to enhance the finishing steps.
To achieve perfect smoothness on the shell surface, the polishing process consists of multiple stages, each involving specialized machinery operated continuously for several hours. After each polishing stage, the fuel capsules are removed, and surface roughness measurements are manually taken \cite{jin2020gaussian} using Keyence VK-X1100 confocal microscope. After the measuring process, the fuel capsules go through a cleaning phase where they are washed, followed by the next polishing phase. 

The adoption of such long-stretch polishing processes is driven by two primary factors: \begin{enumerate*} \item Hard and Brittle Shell Surfaces and \item Unconventional Polishing Process \end{enumerate*}. Regarding the Hard and Brittle Shell Surfaces, the diamond-coated shell surfaces are hard and brittle which makes it essential to avoid high material removal rates during polishing -- excessive material removal could damage the surface and hinder the performance in the experiments. Regarding the Unconventional Polishing Process, the surface polishing technique that is required differs from the techniques used in traditional manufacturing. Naturally, this results in a scarcity of technical knowledge and guidelines for the entire polishing process. Given those reasons we usually choose to overpolish rather than underpolish, which helps the shells to have the right surface roughness, even if the polishing process lasts more than needed.

The research goal of our work is to assure that the polishing process is headed in the right direction and make necessary corrections early in the process. Besides, measuring the surface roughness of the shells is a labor-intensive and time-consuming task. {\em In this paper, we try to answer the following question: Can we automatically predict the surface roughness without going through the expensive measurement phase?} Addressing this questions could lead to potential improvements in the polishing process, its time efficiency and operation optimization, and would promote the research in nuclear fusion.

In this work, we use using machine learning to predict the surface roughness in a faster manner. We treat this problem as a regression task. {\em To the best of our knowledge this is one of the first works that tries to address this part of the polishing process at smaller time scales, especially with the combination of machine learning and signal processing techniques.} 
In particular, we integrate an accelerometer within the polisher to collect vibration data during various polishing stages and measure the corresponding surface roughness using the confocal microscope. We then apply machine learning techniques on this data to train models that can accurately predict the surface roughness. Our models learn critical features in the vibration spectra and correlate them with the surface roughness. However, since measurement of surface roughness is very time-consuming and labor-intensive, we have very limited data: only 24 vibration samples and their corresponding surface roughness values.

We applied several different machine learning techniques that are suitable for regression on a very small dataset: Gaussian Process Model, Linear Regression, Ridge Regression, Regression Decision Tree, Random Forest, Support Vector Regression, and Gradient Boosting Regression. We did not apply deep learning as it is not applicable to such a small amount of data. We found Regression Decision Tree to be the best, showing an average absolute error of 0.42 nm. In future work, we plan to integrate this model with the accelerometer to perform near-sensor data analysis and surface roughness prediction in real time, allowing the human operator to quickly modify the polishing settings if needed for optimal results.


The remainder of the paper is organized as follows. Section 2 provides a background of related works which use machine learning for in-situ process monitoring. In Section 3 we present the polishing experiment details and introduce our methodology. Section 4 presents the numerical experiments pertaining to the different machine learning models. Finally, in section 5, we conclude the paper by summarizing our work and presenting future directions.

\section{Related Work}
\label{sec:related_work}

Several machine learning or data-driven approaches have been proposed for either supervised or unsupervised learning. These works are trying to find a pattern of the input data when a certain output, in our case surface roughness, is desired. \citet{hetherington1999analysis} proposed to use the vibration signal to monitor a continuous layer dielectric wafer surface, while claiming that the attenuation in the vibration intensity is related to the roughness reduction. This point is not valid for nanoscale surface polishing that we need because deciding the threshold for stopping polishing is not trivial.

Two decades later, since \citet{hetherington1999analysis}'s work, the community has moved towards supervised learning setups trying to connect the process conditions and the product surface roughness. In that way, we can anticipate the final roughness with only in-process measurements and further decide when to stop the polishing action. \citet{bukkapatnam2008experimental} used linear and nonlinear regression to correlate the process parameters with the vibration signals to the process condition. Another interesting approach was proposed by \citet{kong2011nonlinear} who used Bayesian models for process state prediction and then used a neural network model to relate the process state prediction with the process condition. In a similar manner \citet{garcia2018multi} deployed feature extraction from the vibration signals by wavelet packet decomposition. Similarly, \citet{plaza2018application} related the vibration data to the surface roughness, with a polynomial regression model.

Much recently, \citet{shilan2023} introduced two different ideas, one in an unsupervised learning setup and the other in a supervised learning setup. Under the unsupervised learning setup using only the process input alone they take advantage of the kinetic energy of vibration signals. Their results show that following only a simple monotonic trend in vibration intensity is not enough. Under the supervised learning setup they built a statistical model that uses limited data and estimates surface roughness and process understanding but is unable to make in-process prediction.

\section{Proposed Methodology}
\label{sec:method}

\begin{figure}[ht!]
\centering
\includegraphics[width=0.25\textwidth]{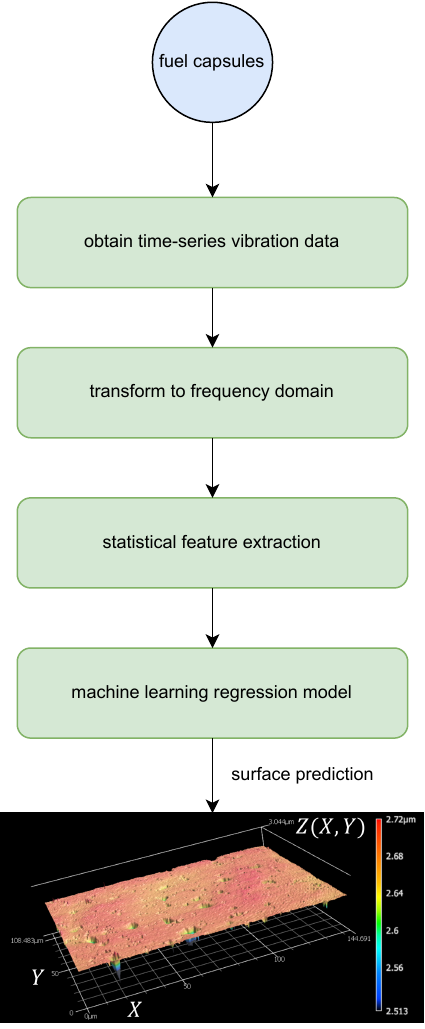}
\caption{Process flow of our proposed methodology. We first obtain the vibration data from the fuel capsules in the time-series domain. We then transform the data to the frequency domain with multiple spectrograms, and we extract statistical features from them. These statistical features are then exploited by a machine learning regression model that predicts the final surface prediction.}
\label{fig:process_flow}
\end{figure}

In this section, we present the polishing experiments, collected data and the methodology to predict polishing surface roughness from the vibration signals. Our methodology has two main steps. In the first step, feature processing of surface measurements and vibration signals are performed. Specifically, roughness from surface measurements and statistical quantifiers of \textit{spectral-bands} energies from time domain vibration signals are determined. In the second step, various machine learning regression models are applied on surface and vibration features as output and predictors, respectively. The methodology is depicted in \cref{fig:process_flow}.

\subsection{Polishing experiments}

During capsule polishing experiments, two types of data are collected. The first data type is the surface measurement collected as micrographs using a confocal microscope. The second data type is the vibration time series signals collected \textit{in-situ} during polishing, at very high sampling rates. The surface measurements contain topography information as a matrix ${\rm Z}(X,Y)$ denoting the height at location $(X,Y)$. Raw vibration signals $y(t)$
are sampled at $10 {\rm kHz}$. In other words, consecutive samples $y(t)$ and $y(t+1)$ are 0.0001 seconds apart. 

Two polishing experiments are conducted during which the surface measurements and vibration signals are collected. 
Each polishing experiment consists of multiple stages ${\rm I}$ with each stage lasting a certain duration ${\rm T}$.
Surface measurements are collected at the end of each stage, while vibration signals are collected during a polishing stage. Below, we describe each of the two experiments.

\begin{figure}[t!]
\centering
    \includegraphics[width=\columnwidth]{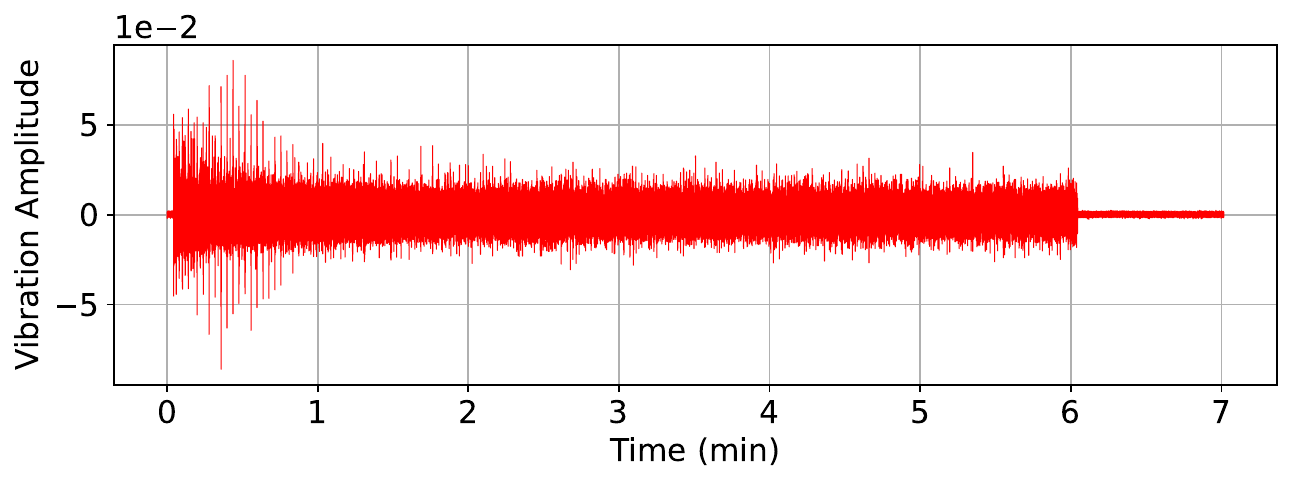}
    \caption{Signal in time domain of one run that is 6 minutes long (Experiment 1)}
\label{fig:example_signal}
\end{figure}

Polishing experiment 1 includes short duration polishing, comprising $I_1 = 18$ stages with each stage comprising ${\rm T_1 =  360\times10^4}$ samples (6 minutes at 10 kHz). Polishing experiment 2, however, are longer duration comprising $I_2 = 6$ stages with each stage lasting for ${\rm T_2 =} 43200\times10^4$ samples (12 hours at 10kHz). In the remainder of the manuscript, superscripts on the collected data are reserved to denote the experiment and stage. For example, $y^{i1}(t)$ and $Z^{i1}(X,Y)$ denote vibration signals and surface measurements collected during stage $i$ of experiment 1 at time $t \in {\rm T}^1$ and end of stage $i$, respectively. The data collected from the polishing experiments are summarized below.

\begin{itemize}[leftmargin=*]
  \item \textbf{Short duration experiments (experiment 1):} \\ $(y^{i1}(t), Z^{i1}(X,Y)) : i \in \{1,\ldots,I_1=18\}, t \in T_1=360\times10^4~samples$
  \item \textbf{Long duration experiments  (experiment 2):}  \\ $(y^{i2}(t), Z^{i2}(X,Y)) : i \in \{1,\ldots,I_2=6\}, t \in T_2 = 4.32\times10^8~samples$
\end{itemize}

For experiment 1 (an example is shown in \cref{fig:example_signal}), we truncate the first and last 1 minute of the time series, which are considered to be transient portions of the data containing undesired data like machine startup and ramp-down. For experiment 2 (which consists of vibration time series for any 12-hour run), we use the last 4 minutes of data, for consistency with the length of data after truncation of experiment 1. 

\subsection{Feature extraction from surface measurements and vibration signals}
\label{subsec:frequ_domain_step_1}

The surface measurements and raw vibration signals undergo additional feature extraction. For elucidation, we avoid superscripts. Specifically, for the surface measurements ${\rm Z}(X,Y)$, a standard metric of surface roughness called the areal surface roughness ($S_a$) \cite{jin2023hypothesis} is derived. This measure of roughness is given by

\begin{equation}
 S_a = \frac{1}{A} \iint_{X,Y} |Z(X,Y) - \overline{Z}|dXdY
\label{eq:ASR}
\end{equation}

\noindent where $A$ is the total area of the micrograph and $\overline{Z}$ is the average height over all $(X,Y)$ locations in matrix ${\rm Z}(X,Y)$. The vibration signals $y(t)$ are transformed into vibration spectra $S_t(f)$ in the time-frequency domain ${\rm T \times F}$ using Short Time Fourier Transform (STFT) \cite{sejdic2009time}. The time-frequency domain reduces the temporal resolution but we obtain information of the energy available in the independent frequencies $f \in {\rm F}$ which can be associated with physical phenomenon during the polishing process. Following this, important spectral-bands $\underline{\alpha}$ associated with changes in the process are identified. Spectral-bands are a subset of the frequency domain  $\underline{\alpha} \subset {\rm F}$. The features used in our study are the spectral-band energies $E(\underline{\alpha})$ given by

\begin{equation}
 E(\underline{\alpha}) = \int_{f\in \underline{\alpha}} S(f)df
\label{eq:SBE}
\end{equation}

\begin{figure}[ht!]
\centering
\includegraphics[width=0.45\textwidth]{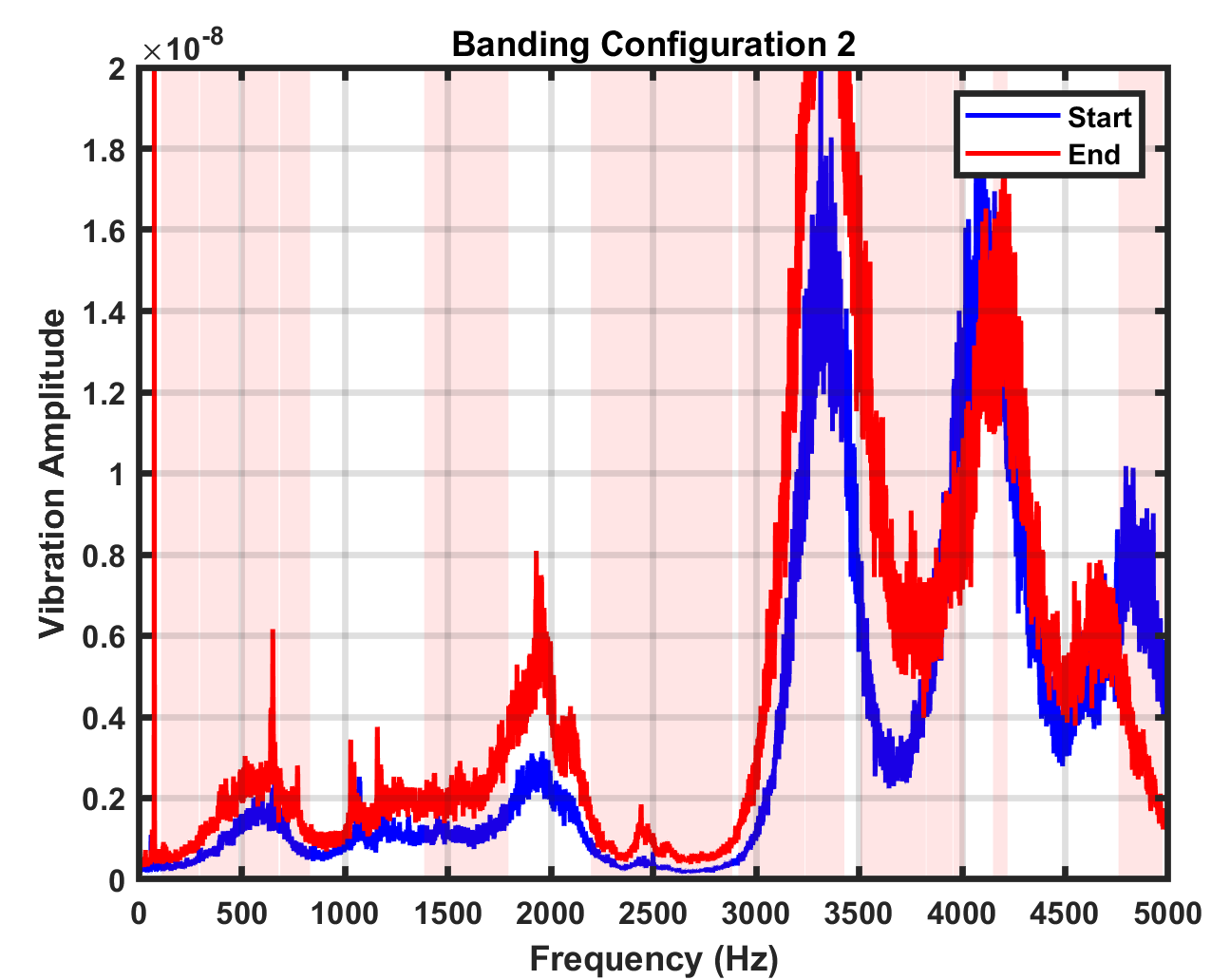}
\caption{Transformed spectrogram that consists of 13 different spectral bands. We observe them with different colors as vertical rectangles in the figure.}
\label{fig:spectrogram}
\end{figure}

During STFT, we use a hamming window size of $1~{\rm second}$ with no overlap between subsequent windows and $2^{14}$ Fast Fourier Transform (FFT) points. This choice for time-frequency domain transformation allows to observe the process in real-time at a time resolution of 1 second and a frequency resolution of ${0.61~\rm Hz}$ with $0 - 5 ~ {\rm kHz}$ range (Nyquist frequency). In our study, we identify 13 spectral-bands ($\underline{\alpha}_1,\ldots,\underline{\alpha}_{13}$) which indicate changes during the process. These spectral-bands are illustrated as red silhouettes in \cref{fig:spectrogram} over laid on top of two spectrums $S_{T_{start}}(f)$ and $S_{T_{end}}(f)$ from the start and end of the polishing process in a polishing stage respectively.


\subsection{Statistical quantifiers for spectral-band predictors}

After the creation of the spectrograms, we extract statistical features from them. These include mean, variance, kurtosis and skewness. The reason for the choice of these statistical features is their popularity and the ease of computation. In more detail, we use the same notation for all four statistical features, so let us have a set of $n$ values $x_1, x_2, \dots, x_n$. The mean, also known as the average, represents the sum of all values in a dataset divided by the number of values. The mean $\mu$ is calculated as $\mu = \frac{1}{n} \sum_{i=1}^{n} x_i$

The variance measures the average of the squared differences between each data point and the mean. It quantifies the spread or dispersion of the data, and is defined as $\sigma^2 = \frac{1}{n} \sum_{i=1}^{n} (x_i - \mu)^2$

Kurtosis is a measure of the "tailedness" of a distribution, indicating the extent to which data values cluster in the tails compared to a normal distribution, and is defined as $K = \frac{\frac{1}{n} \sum_{i=1}^{n} (x_i - \mu)^4}{\left(\frac{1}{n} \sum_{i=1}^{n} (x_i - \mu)^2\right)^2}$

Skewness measures the asymmetry of a distribution, indicating whether the data is skewed to the left (negatively skewed) or to the right (positively skewed) and is defined as $S = \frac{\frac{1}{n} \sum_{i=1}^{n} (x_i - \mu)^3}{\left(\frac{1}{n} \sum_{i=1}^{n} (x_i - \mu)^2\right)^{3/2}}$

\subsection{Machine Learning Techniques}
\label{subsec:machine_learning_techniques_step_2}

After the extraction of the features, we apply several different machine learning techniques that are applicable to very small datasets. The models that we used are: Guassian Process model \cite{rasmussen2006gaussian}, Linear Regression \cite{montgomery2021introduction}, Regression Decision Tree \cite{breiman2017classification}, Ridge Regression model \cite{hastie2009elements}, Random Forest \cite{breiman2001random}, Support Vector Regression \cite{cortes1995support}, and Gradient Boosting Regression \cite{friedman2001greedy}.

\paragraph{\textbf{Gaussian Process Model.}} The Gaussian Process (GP) model is a powerful and flexible probabilistic machine learning approach and is defined as a collection of random variables, any finite number of which have a joint Gaussian distribution. It assumes that any finite set of function values follows a multivariate Gaussian distribution, and the properties of the GP are fully characterized by its mean function and covariance function (also known as kernel). The mean function $\mu(x)$ represents the expected value of the function at input point $x$. It is often assumed to be zero in many cases and is depicted as:

\begin{equation}
    \mu(x) = E[f(x)]
    \label{eq:mean_gp}
\end{equation}

And the covariance function or kernel is $k(x,x^\prime)$ models the correlation between the points $x$ and $x^\prime$. A common choice for that is the squared exponential kernel RBF:

\begin{equation}
    k(x, x') = \exp\left(- \frac{{||x - x'||^2}}{{2 \cdot \text{{length\_scale}}^2}}\right)
    \label{eq:kernel_gp}
\end{equation}

where $\text{{length\_scale}}$ controls the correlation's rate of decrease with distance. The joint distribution of GP is then:

\begin{equation}
\begin{bmatrix}
f \\
f^*
\end{bmatrix}
\sim \mathcal{N}\left(
\begin{bmatrix}
\mu(X) \\
\mu(X^*)
\end{bmatrix},
\begin{bmatrix}
K(X, X) & K(X, X^*) \\
K(X^*, X) & K(X^*, X^*)
\end{bmatrix}\right)
\label{eq:joint_gp}
\end{equation}

GPs are particularly useful when dealing with limited data, providing not only accurate predictions but also robust estimates of uncertainty. Thus, GP makes a perfect candidate for our task. So in our case, this collection of random variables from GP, tries to fit a joint Gaussian distribution to all the four extracted features (Mean/Variance/Kurtosis/Skewness) from the 13 important frequency bands in the frequency domain (extracted in Section 3.2).

\paragraph{\textbf{Linear Regression.}} 

The Linear Regression model represents the relationship between the input extracted features $X$ and the target surface roughness $Y$ as follows:

\begin{equation}
   Y = \beta_0 + \beta_1 X_1 + \beta_2 X_2 + \ldots + \beta_p X_p + \varepsilon
   \label{eq:lr_equation}
\end{equation}

where $Y$ is the target variable, $X_1, X_2, \ldots, X_p$ are the independent variables (features, and in our case the four statistical quantifiers computed on the 13 important frequency bands), $\beta_0, \beta_1, \beta_2, \ldots, \beta_p$ are the coefficients (weights), and $\varepsilon$ is the error term. The objective of linear regression is to minimize the sum of squared residuals (differences between predicted and actual values) as follows:

\begin{equation}
    \text{Minimize} \sum_{i=1}^{n} (y_i - (\beta_0 + \beta_1 x_{i1} + \beta_2 x_{i2} + \ldots + \beta_p x_{ip}))^2
    \label{eq:lr_objective}
\end{equation}
    
Here, $n$ is the number of data points, $y_i$ is the actual target value for the $i$th data point, and $x_{ij}$ is the $j$th feature value for the $i$th data point. For multivariate data the equations can be represented in matrix form $\mathbf{Y} = \mathbf{X} \mathbf{B} + \mathbf{\varepsilon}$, where $\mathbf{Y}$ is the vector of target values, $\mathbf{X}$ is the matrix of feature values, $\mathbf{B}$ is the vector of coefficients, and $\mathbf{\varepsilon}$ is the vector of error terms. Linear Regression seeks to find the optimal coefficients $\mathbf{B}$ that minimize the squared differences between the predicted surface roughness and actual surface roughness in a regression setup. 

\paragraph{\textbf{Ridge Regression.}} Ridge Regression is a variant of linear regression that mitigates multicollinearity and overfitting by adding a regularization term to the objective function. This regularization term penalizes large coefficient values, leading to a more balanced model that better generalizes to new data. So by adding a regularization term in \cref{eq:lr_objective} we have the new objective:

\begin{equation}
\text{Minimize} \sum_{i=1}^{n} (y_i - (\beta_0 + \beta_1 x_{i1} + \beta_2 x_{i2} + \ldots + \beta_p x_{ip}))^2 + \lambda \sum_{j=1}^{p} \beta_j^2       
\end{equation}

where $\lambda$ is the regularization parameter. And the representation of Ridge Regression in matrix form is $\text{Minimize} \ (\mathbf{Y} - \mathbf{X} \mathbf{B})^T (\mathbf{Y} - \mathbf{X} \mathbf{B}) + \lambda \mathbf{B}^T \mathbf{B}$.

\paragraph{\textbf{Regression Decision Tree.}} Regression Decision Tree recursively partitions the space of the extracted features into regions and making predictions based on the average (or another aggregation) of the surface roughness within those regions. Each internal node of the tree represents a decision based on a feature and split point, leading to a hierarchy of decisions that collectively form the prediction. It operates in the following way: At each internal node of the tree, the algorithm chooses one of the extracted features $j$ and a split point $s$ to partition the data into two subsets $R_1(j, s)$ and $R_2(j, s)$:

\begin{equation}
    \begin{split}
   R_1(j, s) = \{x | x_j \leq s\} \\
   R_2(j, s) = \{x | x_j > s\}
   \label{eq:tree_split}
    \end{split}
\end{equation}

The prediction for a leaf node \(R\) is the average (or another aggregation) of the target values \(y_i\) within that region:

\begin{equation}
   \hat{y}_{R} = \frac{1}{N_R} \sum_{i \in R} y_i
   \label{eq:tree_prediction}
\end{equation}

\paragraph{\textbf{Random Forest.}} Random Forest is an ensemble learning method that combines multiple individual decision trees to improve predictive accuracy and reduce overfitting. Random Forests are robust, handle high-dimensional data well, and also provide insights into feature importance. In each tree-building iteration, Random Forest samples the training data with replacement, creating bootstrapped samples $S_b$ of size $N$ from the original dataset of $N$ data points. For each tree, a random subset of features is selected to split nodes during training. This helps to decorrelate the trees and enhance diversity. The final prediction from the Random Forest is the average of predictions from individual trees:

\begin{equation}
    \hat{y} = \frac{1}{B} \sum_{b=1}^{B} \hat{y}_b
    \label{eq:random_forest}
\end{equation}

where $B$ is the number of trees, and $\hat{y}_b$ is the prediction from the $b$th tree.

\paragraph{\textbf{Support Vector Regression.}} Support Vector Regression (SVR) works by finding a hyperplane that best fits the data while allowing a specified margin of error (epsilon). SVR aims to find the hyperplane that minimizes the sum of the errors while penalizing deviations larger than epsilon. It's particularly effective when dealing with non-linear relationships and handling outliers. The objective of SVR is to find the hyperplane $f(x) = \mathbf{w} \cdot \mathbf{x} + b$ that minimizes the following cost function:

\begin{equation}
    \text{Minimize} \ \frac{1}{2} ||\mathbf{w}||^2 + C \sum_{i=1}^{n} \max(0, |y_i - f(x_i)| - \epsilon)
    \label{eq:svr_objective}
\end{equation}

where $||\mathbf{w}||^2$ represents the regularization term, $C$ controls the trade-off between regularization and fitting errors, and $n$ is the number of data points. SVR introduces an epsilon-insensitive tube around the hyperplane. Data points falling within this tube do not contribute to the error, and points outside the tube are penalized according to their distance from the tube boundaries. The prediction for a new data point $x$ is given by $f(x)$, where $f(x)$ is the value predicted by the hyperplane.

\paragraph{\textbf{Gradient Boosting Regression.}} Gradient Boosting Regression is another ensemble learning technique that builds a predictive model by combining the outputs of multiple weak learners like decision trees sequentially. It aims to correct errors made by previous models in an additive manner. Each new model is trained on the residual errors of the combined model so far. Gradient Boosting Regression minimizes the following objective function:

\begin{equation}
   \text{Minimize} \sum_{i=1}^{n} L(y_i, F(x_i)) + \sum_{m=1}^{M} \Omega(f_m)
    \label{eq:gbr:objective}
\end{equation}

where $L$ is the loss function, $F(x_i)$ is the current prediction for data point $x_i$, $M$ is the number of weak learners, and $\Omega(f_m)$ is a regularization term for the $m$th tree. At each iteration, a new weak learner $f_m$ is trained to minimize the negative gradient of the loss function with respect to the current combined model:

\begin{equation}
   -\left[\frac{\partial L(y_i, F(x_i))}{\partial F(x_i)}\right]_{F(x)=F_{m-1}(x)}
    \label{eq:gbr_loss}
\end{equation}

And then the final prediction is obtained by adding the predictions of all weak learners:

\begin{equation}
   \hat{y} = F(x) = F_0(x) + \alpha_1 f_1(x) + \alpha_2 f_2(x) + \ldots + \alpha_M f_M(x)
    \label{eq:gbr_prediction}
\end{equation}

where $F_0(x)$ is the initial prediction, $f_m(x)$ is the prediction of the $m$th weak learner, and $\alpha_m$ is the learning rate for the $m$th iteration.

\section{Numerical Experiments}
\label{sec:experiments}

In this section, we give details on the data and the experiments that we conducted for our study. The polishing vibration data and corresponding surface roughness values that we are using comes from 6-minute runs and 12-hour runs (Section 3.1). The data is limited since we have only 18 6-minute runs and 6 12-hour runs. This is not an obstacle though because the methods that we chose to deploy are suitable for regression tasks with limited data. To properly test the data, we decided to perform a one-fold-cross-validation on the combined data from both the 6-minute runs and the 12-hour runs. This means that for each epoch, we choose 1 run as testing data and the rest of the runs as training data. We repeat this process for the number of runs, which in our case is 24 runs in total and then take an average of the prediction accuracies. As the prediction target for our regression problem, we use the difference between the final surface roughness after each polishing run and the starting surface roughness before that polishing run {\em (i.e., change in surface roughness during a polishing run is our prediction target).} This difference varies from 0.008 to 3.02 for the 6-minute data, and from 1.06 to 2.28 for the 12-hour data. We deployed all the methods from the sklearn library \cite{scikit-learn} which provides effectiveness and simplicity. 

We conducted two sets of experiments. In the first one, we extract features from all the frequency bands together so we will have 1 mean/variance/skewness/kurtosis for each run. In the second experiment, each band gets its own feature, so in our case since we have 13 spectral bands, we will have 13 mean/variance/skewness/kurtosis for each run. We evaluate our approach with the Mean Absolute Error (MAE), which is a well known metric for regression tasks. The MAE measures the average absolute difference between predicted values and actual values. It treats all errors equally regardless of their magnitude. We have a set of ground truth values $y_i$ and the predicted values $f(x_i) \forall i=1,2,\dots,n$ so the MAE is computed as:

\begin{equation}
    MAE = \frac{1}{n}\sum_{i=1}^n|y_i-f(x_i)|
    \label{eq:mae}
\end{equation}

The results of the experiment can be found in \cref{tbl:results}. We observe that the Regression Decision Tree in the case when we use extracted features from each spectral band separately performs the best. In the same case, Linear Regression completely fails when we use extracted features from each spectral band separately. Additionally, the results show that using extracted features from each spectral band separately achieves better accuracy than extracted features from all the bands together. From the other models, we distinguish the good performance of Gradient Boosting Regression, which is close to the Regression Decision Tree, but not better. We also observe that the more involved models like Random Forest, Support Vector Regression, and Gradient Boosting Regression, all perform better than the simpler models, such as Linear Regression, Ridge Regression, Gaussian Process.

To further investigate our approach we checked the feature importance of the best performing method along with the best feature extraction approach. More specifically we checked the feature importance output of the Regression Decision Tree to find out which spectral bands are the most important that made up the final prediction. From our experiments the Regression Decission Tree used only features extracted from the spectral bands 2, 9, and 12. While the model splits in these three spectral bands, the rest of the features are given zero score of importance according to the model. 

\begin{table}[ht]
\centering
\small

\begin{tabular}{|c| *{2}{>{\centering\arraybackslash}p{2cm}|}}
    \hline
    \textbf{Method} & \textbf{Features Together} & \textbf{Separate Features} \\
    \hline 
    \hline
    Linear Regression & 0.91 & 5.5 \\
    \hline
    Gaussian Process & 1.14 & 1.14 \\
    \hline
    Decision Tree & 0.89 & {\bf 0.42} \\
    \hline
    Ridge Regression & 0.74 & 0.73 \\
    \hline
    Random Forest & 1.04 & 0.62 \\
    \hline
    Support Vector & 0.7 & 0.59 \\
    \hline
    Gradient Boosting & 1.32 & 0.46 \\
    \hline
\end{tabular}
\caption{Table depicts the main results of our experiments. The second column "Features Together" shows the MAE when we extract features from all the spectral bands together. The third column "Separate Features" shows the MAE when we extract features from each spectral band separately. The best results are obtained with the Regression Decision Tree when we extract features separately from each spectral band, as shown in bold.}
\label{tbl:results}
\end{table}

\section{Summary and Conclusions}
\label{sec:summary}

In this work, we presented and compared performance of several machine learning models for predicting surface roughness of ICF capsule targets during polishing using vibration signals. The ability to predict surface roughness allows to use vibration signals as surrogates of the state of capsule surface quality in real-time, saving us time and resources in the polishing process of the capsule polishing for nuclear fusion targets. 

Some of the directions our current work can be extended are as follows. After identification of promising machine learning models, multiple models can be used as part of an aggregated committee-of-learners model as an ensemble, which can improve the generalization of our models. Another avenue that we can follow is to look in more detail in the feature extraction process of our proposed methodology. Looking into more involved signal processing techniques could assist us to extract more meaningful features for the machine learning models.

\begin{acks}
This work was performed under the auspices of the U.S. Department of Energy by Lawrence Livermore National Laboratory under Contract DE-AC52-07NA27344 and by the LLNL LDRD program under Project Number 23-ERD-014 (IM: LLNL-CONF-853335).
\end{acks}

\bibliographystyle{ACM-Reference-Format}
\bibliography{bibliography}

\appendix


\end{document}